\documentclass[11pt]{article}

\usepackage[preprint]{acl}
\usepackage{times}
\usepackage{latexsym}
\usepackage[T1]{fontenc}
\usepackage[utf8]{inputenc}
\usepackage{microtype}
\usepackage{inconsolata}
\usepackage{graphicx}
\usepackage{booktabs}
\usepackage{amsmath}
\usepackage{amssymb}
\usepackage{array}
\usepackage{float}

\makeatletter
\renewenvironment{abstract}%
  {\par
   \begingroup
   \setlength{\parskip}{0pt}%
   {\centering\large\textbf{\abstractname}\par}%
   \nobreak\vspace{8.2pt}%
     \setlength{\leftskip}{0.6cm}%
     \setlength{\rightskip}{0.6cm}%
     \setlength{\parindent}{0pt}%
     \@setsize\normalsize{12pt}\xpt\@xpt
     \noindent\ignorespaces
  }%
  {\par\endgroup}
\makeatother

\newcommand{\system}{\textsc{CiteChoice}}
\newcommand{\pp}{\,pp}

\newenvironment{promptcard}[2]{%
  \par\smallskip\noindent\rule{\linewidth}{0.9pt}\par\nobreak\vspace{2pt}%
  \noindent{\small\textbf{#1}}\hfill{\footnotesize\textsc{#2}}\par\nobreak%
  \vspace{2pt}\noindent\rule{\linewidth}{0.4pt}\par\nobreak\vspace{3pt}%
  \footnotesize\ttfamily\raggedright\setlength{\parskip}{2pt}%
}{%
  \par\vspace{3pt}\noindent\rule{\linewidth}{0.9pt}\par\smallskip%
}

\title{CITECHOICE: A Causal Audit of How Document Presentation Redistributes Citation Credit in Agentic Search}

\author{Sriram Selvam \\
  \texttt{selvamsriram@gmail.com} \\\And
  Anneswa Ghosh \\
  \texttt{anneswaghosh@gmail.com} \\}

\hypersetup{
  pdftitle={CITECHOICE: A Causal Audit of How Document Presentation Redistributes Citation Credit in Agentic Search},
  pdfauthor={Sriram Selvam; Anneswa Ghosh}
}

\begin{document}
\maketitle

\begin{abstract}
When several retrieved sources support the
same claim, an answer engine cites some but
not others. We call this decision \emph{citation allocation} and introduce
\system{}, a causal audit of authentic multi-turn agentic search. From 129
everyday-query transcripts, \system{} selects 113 same-call document pairs with
independently verified support for the same pre-specified fact, without
observing ranks or answer outcomes; blinded human review confirms 103. It runs
a hash-verified $2{\times}2$ replay crossing pair order with
jointly generated, fidelity-checked structured and prose renderings of one
target while the rest of the transcript remains fixed.
Three results emerge. First, and most importantly, structured rendering
concentrates citation credit rather than clearly increasing source admission.
It raises target citation count by $+0.50$ citations per answer (95\% CI
$[+0.20,+0.84]$; Holm-adjusted $p{=}.033$), without increasing total citations
or reducing competitor credit. The pre-specified incidence
effect (whether the target is cited at all) is $+4.5\pp$ and inconclusive
(95\% CI $[-1.4,+10.4]$; $p{=}.168$). Second, observational position
differences exceed controlled reordering effects: the
rank-1--rank-5 citation gap is 42.3 points, compared with $+7.9\pp$ in the
main replay and $0.0\pp$ held out. Third, citation evaluation has a measurable
noise floor. Although the aggregate count effect repeats under fresh decoding
of 30 frozen families, 15\% of binary decisions change and decoding accounts
for an estimated 45\% of single-generation family-effect variance. Together,
these findings isolate what survives control: presentation can causally
redistribute visible citation credit within frozen transcripts. They do not
establish reliable source admission, a pure formatting mechanism, or a general
rank advantage.
\end{abstract}

\section{Introduction}
\label{sec:intro}

Answer engines increasingly stand between people and the web. Systems such as
search-augmented chatbots retrieve documents, write an answer, and attach
citations to the claims they make \citep{nakano2021webgpt,lewis2020rag}. Most
evaluation asks whether those citations are \emph{correct}: does the cited page
support the sentence, and is every claim cited
\citep{gao2023alce,liu2023verifiability,schreieder2026survey}? This paper asks
a question that becomes visible only when several retrieved pages could
support the same claim: \emph{how is the credit distributed, and who receives
it?}

We call this decision \textbf{citation allocation}. A citation is a unit of
visibility in generative search, and publishers already optimize content for
generative engines \citep{aggarwal2024geo}. Determining which document
properties causally move citations, and by how much, is therefore both a
scientific question and an ecosystem question.

\begin{figure}[H]
  \centering
  \includegraphics[width=0.98\columnwidth]{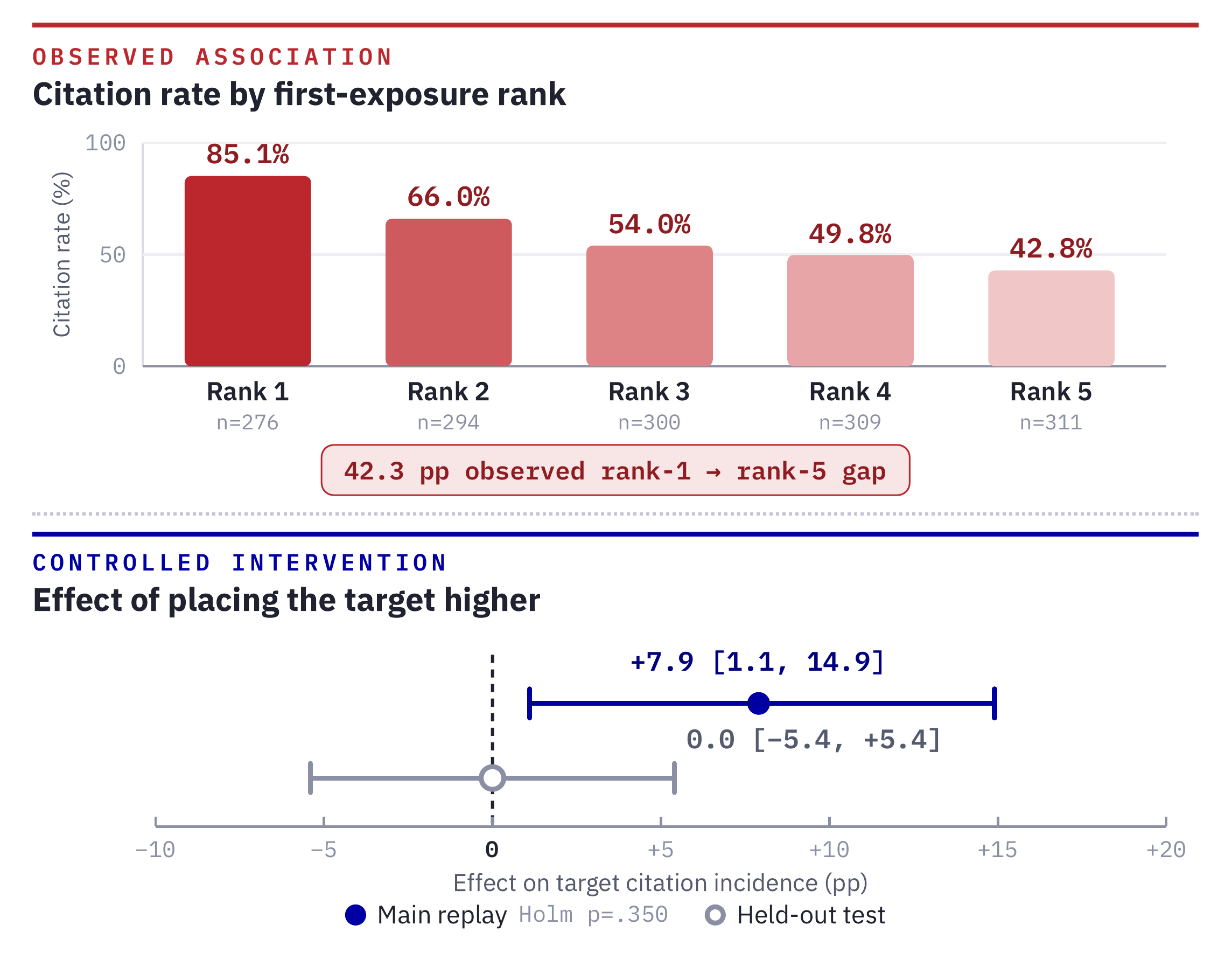}
  \caption{Observed rank gradient (top) and controlled pair-swap effects
  (bottom). These distinct populations and interventions do not form a causal
  decomposition (Section~\ref{sec:rank}).}
  \label{fig:gradient}
\end{figure}

Position makes the problem vivid. Documents first exposed at rank~1 were cited
85.1\% of the time, versus 42.8\% at rank~5. But search engines place more
relevant and higher-quality pages earlier, and the agent chose the queries that
produced those rankings. When we move the \emph{same} document relative to a
matched competitor while holding the transcript fixed, the estimate shrinks
to $+7.9\pp$ in the scaled replay and exactly $0.0\pp$ in a held-out
confirmation (Figure~\ref{fig:gradient}). The contrast is itself a finding:
observational rank gradients can be several times larger than the allocation
change supported by controlled reordering.

\begin{figure*}[t]
  \centering
  \includegraphics[width=0.98\textwidth]{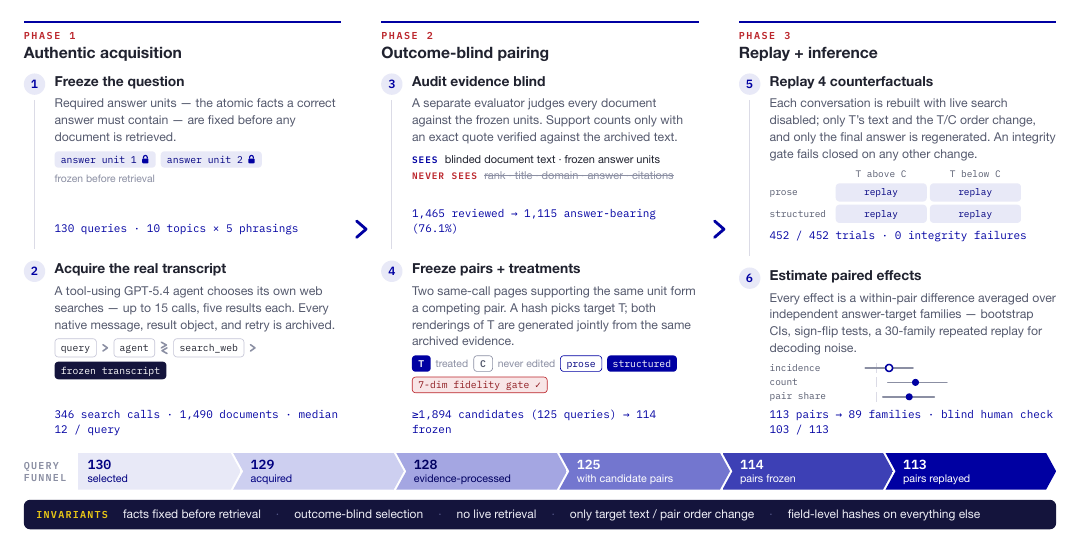}
  \caption{The \system{} protocol from authentic acquisition to paired
  inference. Answer units are fixed before search; evidence review and cohort
  construction are outcome-blind; the replay crosses matched target rendering
  with pair order while hash-checking every other model-visible field.
  Appendix~\ref{app:walkthrough} follows one real query end to end.}
  \label{fig:pipeline}
\end{figure*}

Realistic intervention requires preserving the agent's own search process.
\system{} therefore combines authentic acquisition with counterfactual replay
(Figure~\ref{fig:pipeline}). A tool-using GPT-5.4 agent answers 130 human-style
queries and issues its own web searches; every native message and result object
is archived. We then reconstruct each frozen conversation, change exactly one
declared cue, either the within-call order of two competing documents or the
target's rendering, and regenerate only the final answer. The transcript is
otherwise byte-identical and hash-verified.

The intervention acts on the serialized source snapshot the answer model
actually reads, downstream of retrieval and extraction. This is not an exotic
cue class: among 1,750 archived result-text records, 93.2\% contain
markdown-style headings, 76.6\% contain list-marker lines, and 21.3\% contain
table-like rows. Extraction and serialization therefore decide which document
structure survives into model context; our results show that this boundary can
have downstream consequences for citation credit. Visual layout, typography,
DOM structure, and the live web remain untouched.

The unit of study is a \textbf{competing pair}: two documents returned in the
same search call, each with independently verified evidence for the same
required answer fact. Selection is outcome-blind: code and reviewers never
see ranks, the baseline answer, or citation outcomes. The cohort is therefore
not chosen for movable-looking citations. From at least 1,894 candidates we
freeze one machine-screened pair per query for 114 queries; 113 enter a
$2{\times}2$ replay crossing pairwise order with target rendering, for 452
trials over 89 independent answer-target families. A later blinded human
adjudication confirms 103 of the 113 as strict same-proposition competitions;
all headline estimates strengthen on that subset.

The paper contributes three affirmative findings and one reusable instrument:
\begin{enumerate}
  \setlength{\itemsep}{2pt}
  \setlength{\parskip}{0pt}
  \setlength{\parsep}{0pt}
  \setlength{\topsep}{3pt}
  \item \textbf{Structured rendering concentrates citation credit.} It raises
  target citation count by $+0.50$ without increasing total answer citations;
  the matched competitor is unchanged, and about half the gain attaches to
  shared evidence.
  \item \textbf{Controlled rank effects are sharply smaller than the
  observational gradient.} The 42.3-point descriptive gap contracts to
  $+7.9\pp$ in the scaled swap and $0.0\pp$ held out.
  \item \textbf{Citation evaluation carries a measurable noise budget.} One in
  seven binary decisions flips across identical generations, and 45\% of
  single-draw effect variance is decoding noise.
\end{enumerate}
Beyond those estimates, \system{} contributes a fully traced, outcome-blind,
hash-verified protocol and analysis-reproducible artifact.

What remains open is equally specific. The pre-specified incidence estimate is
$+4.5\pp$ but inconclusive; at the observed variance, 89 families provide
roughly 80\% power only for effects around $8.5\pp$ or larger. The
word-preserving formatting ablation and the general rank mechanism are also
unstable across repeatability or held-out checks. These are limits on mechanism
claims, not negations of the supported concentration, deflation, and noise
findings.

\section{Setting and Terminology}
\label{sec:terms}

The study rests on one construction. An \textbf{answer unit} is an atomic
required fact, fixed before any document is inspected; a document is
\textbf{answer-bearing} for a unit only if an evaluator finds an exact
supporting quote in its archived text. Two same-call documents both
answer-bearing for the same unit form a \textbf{competing pair}: either could
legitimately be cited, so whichever the engine credits is an allocation
decision rather than a correctness decision. Table~\ref{tab:terms} defines the
outcomes and grouping unit used throughout.

\begin{table}[t]
\centering
\footnotesize
\begin{tabular}{@{}p{0.30\columnwidth}p{0.63\columnwidth}@{}}
\toprule
Term & Meaning \\
\midrule
answer unit & atomic required fact, frozen before document review \\
competing pair & two same-call documents sharing verified support for a unit \\
target / competitor & hash-chosen pair member that receives treatments / the never-edited member \\
citation incidence & target cited at least once in the answer (source admission; $Y{\in}\{0,1\}$) \\
citation count & number of valid citation markers assigned to the target \\
pair citation share & target markers divided by target-plus-competitor markers when the pair is cited \\
answer-target family & queries built on the same underlying fact; the independent clustering unit \\
\bottomrule
\end{tabular}
\caption{Terms used throughout. After this definition, ``incidence'' means
cited at least once and ``count'' means the number of target markers. ``pp''
denotes percentage points.}
\label{tab:terms}
\end{table}

\section{Related Work}
\label{sec:related}

\paragraph{Cited generation and its evaluation.}
WebGPT demonstrated browsing agents that answer with references
\citep{nakano2021webgpt}; retrieval-augmented generation made evidence
conditioning standard \citep{lewis2020rag}; ALCE formalized
citation-quality evaluation \citep{gao2023alce}; audits of deployed engines
found frequent unsupported citations \citep{liu2023verifiability}; a recent
survey consolidates the space \citep{schreieder2026survey}; and correctness
is not faithfulness: a marker can be attached without causal reliance on
the source \citep{wallat2024faithfulness}. All of this treats citation as
\emph{verification}. We hold verification fixed by construction and study
\emph{allocation}: visible credit, not reliance.

\paragraph{Position and context use.}
Long-context models use evidence unevenly by position \citep{liu2024lost}
and are distracted by irrelevant context \citep{shi2023distracted}, yet in
realistic RAG mixtures positional effects are smaller than synthetic
benchmarks suggest \citep{cuconasu2025position,hagstrom2025reality}; we
extend this ``smaller than it looks'' pattern from accuracy to citation
allocation. Adversarial work manipulates conversational-search rankings
through page content \citep{pfrommer2024ranking}; we measure the benign
counterpart.

\paragraph{Evidence presentation and provenance.}
Under \emph{conflicting} evidence, models are swayed by relevance and style
\citep{wan2024convincing}, metadata and appearance
\citep{chiang2024metadata}, authorship labels
\citep{abolghasemi2025attribution}, and source type
\citep{schuster2026source}. Our setting differs in regime and outcome: the
paired documents \emph{agree}, and what moves is the inline citation. GEO
\citep{aggarwal2024geo} showed that content edits can raise generative-engine
visibility. FeatGEO optimizes interpretable structural, content, and
linguistic features across engines \citep{liu2026featgeo}; related systems
engineer structural features or diagnose and repair page-level citation
failures \citep{yu2026geosfe,tian2026agentgeo}. These are optimization
studies over constructed or live pages, typically measuring aggregate
visibility. \system{} instead treats presentation as an audit variable: it
freezes the retrieval interface and estimates within-pair allocation between
verified competitors with declared uncertainty.

Closest to our question, \citet{vishwakarma2026whatgetscited} study which of
two competing sources an answer engine cites first, via large factorial
sweeps over constructed two-document RAG contexts across six models; they
find position and relevance dominate while formatting-only changes matter
little. \system{} differs in substrate and outcome: our competitions arise
\emph{naturally} inside frozen multi-turn agentic transcripts with real page
snapshots, pairs are selected outcome-blind on verified shared evidence, and
we measure incidence, count, and share rather than first-citation preference.
The difference is informative: formatting moved little in their synthetic
two-source setting but moves citation count here, so presentation effects
appear to depend on context realism, competition density, and outcome
definition. Contemporaneous observational work likewise associates
structure with citation influence \citep{zhang2026absorption}; we provide a
causal audit of the allocation side without claiming that visible citations
reveal internal evidence reliance.

\paragraph{What is new.}
\system{} is, to our knowledge, the first to combine authentic multi-turn
agentic acquisition, outcome-blind construction of naturally occurring
shared-evidence competitions, and hash-verified counterfactual replay with
paired, clustered inference over citation allocation.
\section{The \system{} Protocol}
\label{sec:protocol}

\subsection{Authentic acquisition}
\label{sec:acquisition}

The acquisition frame is a balanced sample of 130 queries from a manually
revised 500-query set: 13 from each of ten everyday topics and 26 from each of
five phrasings, from neutral fact bundles to terse search-style input
(Appendix~\ref{app:acquisition}), reflecting what people type into answer
engines rather than benchmark trivia.

The GPT-5.4 agent must search before answering. Within five model turns it may
issue up to three parallel \texttt{search\_web} calls per turn (15 total),
inspect results, and search again. Each call returns five Exa results as native
tool messages with up to 10{,}000 characters of page text plus metadata. The
agent cites opaque \texttt{[[cite:SOURCE]]} handles (contract verbatim as
prompt~D.1). It completed 129 of 130 queries and issued 346 completed search
calls over 1{,}490 unique documents (median 12 per query).\footnote{The three
large document totals count different units: 1,490 deduplicated documents;
1,750 serialized result-text records in the acquisition archive, where repeated
exposures are retained; and 2,010 result objects in the broader schema audit
spanning acquisition and replay payloads.} Answers carry a median of 29
citation markers over six distinct documents. The funnel is 130 selected
queries $\to$ 129 acquired $\to$ 128 evidence-processed $\to$ 125 with
candidate pairs $\to$ 114 frozen pairs $\to$ 113 replayed pairs. Every prompt,
native request and response, retry, latency, and token count is archived
(Appendix~\ref{app:acquisition}).

\subsection{From documents to competing pairs}
\label{sec:pairs}

A separate Grok evaluator judges each of 1{,}465 reviewed documents
against its query's frozen answer units, seeing only blinded document
text. The evaluator sees no titles, domains, ranks, answers, or citations.
Support counts only with an exact quote verified against the archived text; absent or
paraphrased quotes are downgraded (prompt~D.3). This yields 1{,}115
answer-bearing documents (76.1\%).

Two documents form a candidate competing pair if they appear in the same
search call, share at least one supported unit, do not materially contradict
it, and are not duplicates or syndicated copies. The acquisition yields at
least 1{,}894 unique candidate pairs across 125 queries (a 30-pair-per-query
extraction cap saturated for 13 queries). The abundance matters: it lets us
apply strict causal screening without selecting on baseline citation behavior.

\subsection{The outcome-blind cohort}
\label{sec:cohort}

Automated screening then requires a unique within-call occurrence,
complete metadata, rewrite-suitable text, comparable answer-unit coverage,
and a bounded length ratio (all eight criteria and the deterministic
one-pair-per-query rule in Appendix~\ref{app:pairs}). We froze
\textbf{one machine-screened pair per query} for 114 queries; a blinded
reviewer audited all 117 proposed candidates, establishing non-duplication
and eligibility for 114. Reviewers and selection code never observed
ranks, baseline answers, or citations, so the estimand covers
\emph{eligible, machine-screened pairs}, rather than pairs picked because a
baseline citation looked movable. A blinded human adjudication later confirmed 103 of the 113
analyzed pairs (91.2\%) as strict same-proposition competitions; the ten
invalid conflate topical relatedness with shared support, and excluding
them strengthens every headline estimate (Appendix~\ref{app:pairs}). One
pair member becomes the \textbf{target} by an outcome-blind hash; its
competitor is never edited.

\subsection{Matched text treatments}
\label{sec:treatments}

\begin{figure*}[t]
  \centering
  \includegraphics[width=0.96\textwidth]{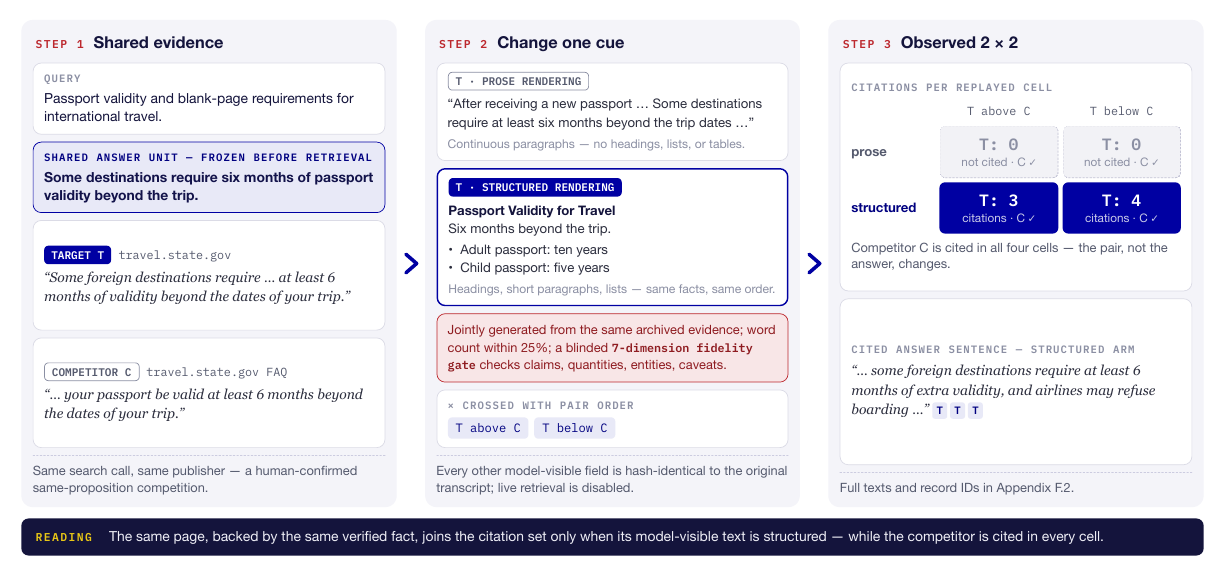}
  \caption{A real, human-confirmed same-publisher pair. Both pages carry
  near-paraphrase support for the six-month passport-validity rule. Only the
  target rendering and pair order change; every other field is hash-identical.
  The target receives 0 citations under prose and 3/4 under structured
  rendering, while the competitor is cited in all four cells. Full texts and
  IDs appear in Appendix~\ref{app:example-structure}.}
  \label{fig:design}
\end{figure*}

The structure intervention asks: \emph{holding the fact inventory fixed,
does organization win citations?} For each target we jointly generate two
renderings from the same archived evidence (Figure~\ref{fig:design}):
\textbf{polished prose} (continuous paragraphs; no headings, lists, or
tables) and \textbf{polished structured} text (headings, short paragraphs,
lists or a table). Both must preserve claims, quantities, entities, caveats, attribution, and
answer-unit coverage, and stay within 25\% in word count. Deterministic
validators enforce length and markup rules, and a separate blinded Grok pass
must pass seven fidelity dimensions (prompts~D.4--D.5). Because \emph{both}
arms are rewrites, the contrast controls for editorial polish. However, the
arms are not token-identical (Jaccard .80; 36\% of structured sentences
verbatim in prose), so the estimand is strictly \emph{jointly generated
structured versus prose rendering}, a rewrite package including wording
changes. To isolate organization
itself, we add a \textbf{mechanical reformat ablation}: the prose arm's exact
word sequence deterministically re-laid-out as one sentence per list row (no
model involved, fidelity guaranteed by construction), replayed over all 113
pairs (Section~\ref{sec:structure}).

Treatment generation succeeded for 113 of 114 targets; the single failure
(a page that repeatedly triggered a provider content filter) was excluded
\emph{before} any scaled outcome existed. A deterministic, topic-stratified,
AI-assisted spot audit reviewed 23
of the 113 accepted treatments and flagged no fidelity violations; it was
not an independent human annotation, and per-item records are archived in
the artifact. Structured variants average only 1.5 words fewer than prose.

\subsection{Frozen counterfactual replay}
\label{sec:replay}

Replay reconstructs the complete provider-native conversation, including the
query, tool calls, tool results, and accumulated result objects, with live
search disabled and deterministic opaque handles replacing ordinal source
IDs. The \emph{rank} intervention exchanges the two pair members' positions
inside their original call, reaching the model purely as array position, with
no numeric rank or score; the \emph{text} intervention swaps in one rendering
of the target. Before any model call, an integrity gate diffs the
manipulated transcript against baseline and fails closed if any undeclared
field changed. All 452 trials passed; each archives pre- and
post-intervention hashes. Appendix~\ref{app:config} lists every
model-visible field.

GPT-5.4 (the acquisition model) generates all primary outcomes: 452 of 452
planned trials completed with no repair, no live search, and no missing
cell (31 transient transport failures across 25 pairs were retried under
unchanged trial definitions).

\section{Estimands and Inference}
\label{sec:inference}

Each pair contributes four trials, one per cell of the $2{\times}2$, so
every effect below is a difference taken \emph{inside} a single pair. For
pair $i$, text arm $c\in\{P,S\}$ (prose, structured), and assigned rank
$r\in\{H,L\}$ (target higher, lower), let $Y_{icr}=1$ if the target is
cited at least once. The pair-level structure effect averages over rank,
$\Delta_i^{\text{struct}}=\tfrac{1}{2}(Y_{iSH}+Y_{iSL})-\tfrac{1}{2}(Y_{iPH}+Y_{iPL})$;
the rank effect averages over text arms analogously, and the interaction is
the difference between the rank effects under structured and prose text.

\paragraph{Estimand.} Answer generation is stochastic and the primary
design draws one generation per cell: our estimand is the \emph{expected}
effect over decoding randomness, conditional on the frozen transcripts,
cohort, and answer-engine configuration. It is not a claim about other
retrievals or queries. Because the whole answer is regenerated under each
treatment, every effect is an \emph{end-to-end source-visibility} effect:
uptake, phrasing, and sentence boundaries can move along with source choice; the
channels are separated only post hoc (Section~\ref{sec:allocation}).
Generation noise is absorbed into the family-level variance the resampling
estimates (valid, conservative in power; Appendix~\ref{app:repeat}
decomposes the noise share).

Some queries were built around the same underlying fact, so their pairs are
not independent: we average pair effects within each \textbf{answer-target
family} and define all estimands over the 89 independent families behind the
113 pairs, reporting pair-level results as sensitivity analyses. Confidence
intervals use a 20{,}000-replicate family bootstrap; two-sided $p$-values
use a sign-flip permutation test on nonzero family effects (exact up to 20
discordant families, 200{,}000 seeded flips beyond). The single primary
hypothesis, declared before scaled outcomes existed, is the structure
incidence effect at $\alpha=.05$.
Sixteen declared secondary tests form one Holm-corrected family; everything
else is labeled exploratory, supplementary, or post hoc. Bootstrap
intervals and sign-flip $p$-values are distinct procedures and need not
agree near the boundary; an interval can exclude zero while $p{>}.05$.
Appendix~\ref{app:estimators} states the estimators in full.

\section{Results}
\label{sec:results}

\begin{figure*}[t]
  \centering
  \begin{minipage}[t]{0.49\textwidth}
    \centering
    \textbf{A. Target citation-count effects}\\[2pt]
    \includegraphics[width=\linewidth]{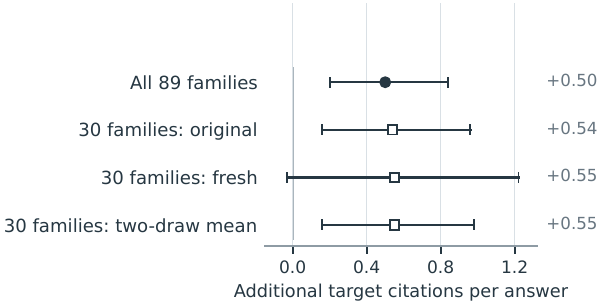}
  \end{minipage}\hfill
  \begin{minipage}[t]{0.49\textwidth}
    \centering
    \textbf{B. Incidence and mechanism checks}\\[2pt]
    \includegraphics[width=\linewidth]{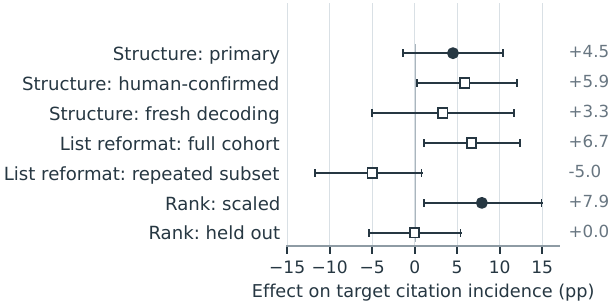}
  \end{minipage}
  \caption{Effect sizes with family-bootstrap 95\% intervals. Panel A shows
  the multiplicity-controlled count effect and its near-identical magnitude
  under fresh decoding of the same 30 frozen families. Panel B scopes what is
  not established: the pre-specified incidence interval crosses zero, the
  word-preserving estimate reverses in the repeated subset, and the scaled
  rank estimate becomes zero held out.}
  \label{fig:effects}
\end{figure*}

The results follow the three claims from the introduction: credit
concentration, rank deflation, and the evaluation noise budget. Mechanism and
cross-model checks then mark the boundaries. Table~\ref{tab:core} reports the
core estimands; Figure~\ref{fig:effects} separates the supported count result
from incidence and mechanism diagnostics.

\begin{table*}[t]
\centering
\footnotesize
\begin{tabular}{@{}p{0.32\textwidth}p{0.16\textwidth}p{0.09\textwidth}p{0.21\textwidth}p{0.11\textwidth}@{}}
\toprule
Estimand & Evidence role & $N$ & Effect (95\% CI) & Test \\
\midrule
Target citation count: structured vs. prose & secondary & 89/113 & $+0.50$ $[+0.20,+0.84]$ & Holm $p{=}.033$ \\
Target incidence: structured vs. prose & \textbf{primary} & 89/113 & $+4.5\pp$ $[-1.4,+10.4]$ & $p{=}.168$ \\
\quad Human-confirmed pairs only & validated sensitivity & 81/103 & $+5.9\pp$ $[+0.3,+12.0]$ & $p{=}.066$ \\
Shared-evidence target citations: structured vs. prose & post-hoc LLM audit & 89/113 & $+0.25$ markers & $p{=}.023$ \\
Target incidence: higher vs. lower rank, scaled & secondary & 89/113 & $+7.9\pp$ $[+1.1,+14.9]$ & Holm $p{=}.350$ \\
Target incidence: higher vs. lower rank, held out & confirmation & 56/56 & $0.0\pp$ $[-5.4,+5.4]$ & $p{=}1.00$ \\
\bottomrule
\end{tabular}
\caption{Core estimates. $N$ reports answer-target families/pairs. The
family-weighted citation-count effect is the clean allocation result that
survives multiplicity correction. Incidence remains uncertain; the rank
estimate is much smaller than the observed gradient and fails held-out
confirmation.}
\label{tab:core}
\end{table*}

\subsection{Structured rendering concentrates citation credit}
\label{sec:allocation}

Structured rendering changes the amount of credit assigned to the target more
clearly than it changes whether the target enters the citation set. Raw
trial-level means are 2.69 target markers per answer under prose and 3.12 under
structured rendering, a 16\% increase. The family-weighted causal estimand is
$+0.50$ citations (95\% CI $[+0.20,+0.84]$; raw $p{=}.0022$, Holm
$p{=}.033$), one of two secondary tests surviving correction. Distinct
answer sentences citing the target rise by the same $+0.50$, so the result is
not repeated markers on one sentence.

The answer-wide citation budget does not expand: structured answers contain
$0.49$ fewer total markers, $0.05$ fewer unique cited sources, and 42 fewer
characters on average, none reliably nonzero. The matched competitor changes
by only $-0.02$ markers ($[-0.28,+0.24]$). Among the subset with pair citations
in both rank cells, the higher-ranked target's within-pair share rises by
$+10.5\pp$ under structured rendering (Holm $p{=}.0018$), although that
72-family population is selected on realized citations and is therefore
descriptive. The count estimand is cleaner.

The pre-specified incidence effect is positive but unresolved: $+4.5\pp$
(95\% CI $[-1.4,+10.4]$; $p{=}.168$), with 18 families favoring structured
rendering, 10 favoring prose, and 61 unchanged. At the observed variance, the
89-family design has roughly 80\% power only for incidence effects around
$8.5\pp$ or larger. The human-confirmed 103-pair subset is $+5.9\pp$
($p{=}.066$), and record-disjoint, cross-model, and fresh-decoding checks are
all directionally positive (Appendix~\ref{app:secondary}); none converts the
primary estimate into a confirmed admission effect.

The supported count magnitude is repeatable at the aggregate level. On 30
hash-chosen frozen families it is $+0.54$ in the first generation, $+0.55$ in
a fresh one, and $+0.55$ after averaging both ($p{=}.013$). This is
independent decoding of the same transcripts, not an independent-dataset
replication.

\subsection{Where the extra credit comes from}
\label{sec:where-credit}

\looseness=-1
The added credit attaches in substantial part to evidence already in
competition rather than to new answer content. A post-hoc blinded LLM audit
aligns all 1,313 target markers to the pair's verified shared units: 48.6\%
attach to shared evidence, 50.7\% to target-specific or other content, and
0.7\% are unclear. Structured rendering adds $+0.25$ shared-evidence markers
($p{=}.023$), which is half the total count effect, and raises the probability
of any shared-evidence citation by $+8.7\pp$ ($p{=}.0005$; both uncorrected
and post hoc).
A second blinded LLM audit finds shared-unit expression nearly unchanged
($+0.8\pp$, 95\% CI $[-1.7,+3.2]$) while target credit rises $+3.6\pp$
($p{=}.042$), with no competitor loss. Point estimates instead place the
offset among other sources ($-0.97$ markers, $[-1.98,+0.01]$, $p{=}.058$).
Thus the answer-level pattern is concentration: the target receives more
credit for evidence the answer already expresses, while the matched rival
remains present. Figure~\ref{fig:design} shows this behavior in a
human-confirmed same-publisher pair; the audits remain post hoc and do not
identify internal evidence reliance.

\subsection{Rank: the observed gradient overstates the controlled effect}
\label{sec:rank}

The observational rank association is much larger than the effect supported
by controlled reordering. First-exposure citation incidence falls 42.3 points
from rank~1 to rank~5. In the matched replay, placing the target above its
competitor changes incidence by $+7.9\pp$ (95\% CI $[+1.1,+14.9]$; raw
$p{=}.035$, Holm $p{=}.350$); in 56 held-out pairs swapped in isolation, the
estimate is exactly $0.0\pp$ ($[-5.4,+5.4]$). The quantities differ in
population, text regime, and channel, so Figure~\ref{fig:gradient} is not a
causal decomposition. It is a deflationary result: the descriptive gradient
is five times the scaled point estimate and has no counterpart in held-out
confirmation.

The scaled estimate is heterogeneous by displacement: exploratory values are
$-3.3\pp$ at one slot, $+13.4$ and $+13.0\pp$ at two and three, and
$+30.0\pp$ for ten full-window swaps. Different pairs occupy each stratum, so
this is compositional rather than an identified dose-response. The held-out
pairs cover comparable one-to-four-slot swaps and remain near zero in every
stratum, including nine full-window cases (Appendix~\ref{app:geometry}).

Replay payloads contain no provider score or numeric rank field; order reaches
the model only through array position. The evidence therefore supports a
narrow conclusion: message position can move attribution in some frozen
transcripts, especially in wider swaps, but average rank effects inferred from
observational gradients are not reliable. This extends the realistic-RAG
``smaller than it looks'' pattern from answer quality to citation allocation
\citep{cuconasu2025position,hagstrom2025reality}.

\subsection{Mechanism checks rule out simple stories}
\label{sec:structure}

Neither list markers nor array position provides a stable one-variable
mechanism. The post-hoc word-preserving ablation compares prose with the exact
same word sequence laid out one sentence per list row. Over all 113 pairs it
raises incidence by $+6.7\pp$ (95\% CI $[+1.1,+12.4]$; $p{=}.028$), but on
the 30 repeatability families it is $-1.7\pp$ initially and $-8.3\pp$ fresh
($-5.0\pp$ averaged, $p{=}.22$). The full-cohort result suggests
serialization can matter; the reversal precludes a stable list-marker
mechanism.

Likewise, the rank-by-structure interaction is $-4.5\pp$ ($p{=}.418$), and
the held-out average rank effect is zero. These boundaries rule out a simple
optimization recipe in which adding lists or moving one source upward
reliably wins citations across transcripts. The supported claim is broader
and end to end: model-visible presentation redistributes credit, while the
operative pathway remains transcript-dependent.

\subsection{Citation evaluation has a noise budget}
\label{sec:stability}

A single generated answer is a noisy measurement of citation allocation. In
120 independently regenerated cells, binary target citation agrees 85.0\% of
the time, meaning one in seven decisions flips. Exact cited-pair sets agree
78.3\%, and exact target counts only 61.7\%. Exact family effects agree for 17
of 30 families even though the aggregate count magnitude repeats.

A method-of-moments decomposition attributes 45\% of single-draw family-effect
variance to decoding noise, with the rest reflecting between-family
heterogeneity (Appendix~\ref{app:repeat}). Repeating $k$ generations per cell
would shrink that noise component by approximately $1/k$. Citation studies
that report one generation per condition therefore inherit a quantifiable
noise floor rather than merely unspecified stochasticity.

\subsection{Across waves and models}
\label{sec:generalization}

The earlier 100-query wave is record-disjoint but not fully independent: 49
answer-target families recur under different queries. Its aggregate direction
is positive, but recurring-family effects correlate negatively
($r{=}-.445$; only four are nonzero in both waves), indicating
transcript-specific effects.

Grok~4.3 replay of the same trajectories is directionally positive after
repair ($+5.6\pp$), yet only 44.7\% of initial responses satisfy the citation
contract; the one-shot estimate is $+0.9\pp$, GPT--Grok agreement is 69.9\%,
and worst-case completion of five missing cells bounds the repaired effect at
$+2.7$ to $+7.1\pp$ (Appendix~\ref{app:missingness}). This supports direction
in one additional model configuration, not model-independent magnitude.

\section{Discussion}
\label{sec:discussion}

\noindent\textbf{For citation evaluators.}
Single answers are noisy: 15\% of binary cells flip under identical
regeneration, and decoding contributes 45\% of single-draw effect variance.
Evaluations should repeat cells and report agreement, between-generation
variance, and aggregate versus per-example stability.

\noindent\textbf{For auditors and publishers.}
Presentation can move visible credit among evidence-matched sources, but the
supported effect is concentration, not admission: the competitor is unchanged,
and no stable list-marker or rank recipe emerges. This is an
attribution-sensitivity warning, not an optimization tactic.

\noindent\textbf{For Document AI pipelines.}
Extraction preserves headings, lists, and table-like rows in model-visible
text; serializers and extractors are therefore part of the attribution
pipeline and should be versioned and audited. The outcome is visible credit,
not causal evidence reliance \citep{wallat2024faithfulness}.

\section{Conclusion}
\label{sec:conclusion}

\system{} causally audits citation allocation through frozen transcripts,
outcome-blind shared-evidence competitions, and hash-verified replay.
Structured rendering concentrates credit without expanding the citation
budget; observational rank gradients exceed controlled effects; and
single-generation evaluation is noisy. Source admission, pure serialization,
and a general rank mechanism remain open; confirmation should make count and
incidence co-primary, repeat cells, harmonize the citation contract, and vary
fresh trajectories, retrievers, and models. Presentation shapes attribution.

\section*{Limitations}

\paragraph{External validity.}
All estimates are conditional on one retrieval provider (Exa, five results per
call, bounded text), one acquisition agent and primary answer deployment
(GPT-5.4), and the authentic transcripts it produced. Frozen replay cannot
measure whether a publisher-side edit changes retrieval, subsequent search
turns, or live-web ranking. The Grok check reuses GPT-acquired trajectories and
requires protocol repair for most responses. Closed deployments can drift
despite archived identifiers and parameters.

\paragraph{Power and multiplicity.}
The main analysis contains 89 independent answer-target families and only 28
discordant families on the primary incidence outcome. Under the observed
variance, the approximate 80\% minimum detectable incidence effect is
$8.5\pp$; the $+4.5\pp$ estimate is therefore unresolved rather than evidence
of zero. The citation-count result survives correction but is one member of a
declared 16-test secondary family.

\paragraph{Construct and treatment.}
The primary contrast is jointly generated structured versus prose rendering,
a rewrite package with lexical differences. No cell in the scaled $2{\times}2$
uses the untouched original target text; both arms also compress the archived
target symmetrically. The mechanical ablation isolates only one serialization
form (sentence-per-list-row), not headed sections, tables, visual layout, DOM
structure, or publisher-side pages. The strict citation contract (median 29
markers per answer) may not transfer to systems with sparse citations, and the
outcome measures visible credit rather than reader traffic or causal evidence
reliance.

\paragraph{Mechanism and post-hoc analysis.}
The word-preserving estimate is positive in the full cohort but negative in
the repeated subset; displacement strata are exploratory and compositional;
and the shared-unit alignment and unit-expression analyses are post hoc and
LLM-scored. Because the whole answer is regenerated, estimated effects can
combine changes in evidence uptake, wording, sentence boundaries, and source
attachment.

\paragraph{LLMs in the measurement loop.}
Grok-4.3 generates the treatment renderings and, in a separate blinded pass,
also performs their fidelity gate; this is procedural separation, not an
independent model family. The 23-item treatment spot audit was AI-assisted,
not independent human annotation. Shared-evidence eligibility was produced by
an LLM with exact-quote checks and then adjudicated by a blinded human for
every analyzed pair (103/113 confirmed); the ten invalid pairs show that
semantic relatedness can be mistaken for same-proposition support, although
excluding them strengthens the headline estimates. Treatment fidelity,
outcome scoring, citation alignment, and unit-expression judgments retain
model-dependent error.

\paragraph{Data and reproducibility.}
We archive search-returned snapshots rather than complete webpages. Licensed
third-party text and raw provider traces cannot be redistributed automatically,
so the artifact cannot provide a fully self-contained copy of every source
snapshot. For reproducibility and independent audit, the public artifact at
\url{https://github.com/selvamsriram/CiteChoice} releases the code,
prompts, complete end-to-end pipeline, hashes, derived records, and no-network
analysis regeneration.

\section*{Ethical Considerations}

The study uses synthetic, manually revised queries and public web-search
responses; no private user data or human participants are involved.
Manipulations occur offline in replay and never modify webpages or search
indexes. Because the findings could encourage publishers to optimize
presentation for citations without improving evidence quality, we report
effect sizes with their uncertainty and frame them as audit diagnostics, not
optimization guidance. Third-party page text and provider traces require
licensing review before redistribution; derived tables, code, prompts, and
hashes can be shared.

AI assistance was used to generate the query sets and implement the pipeline.
After the author prepared the first manuscript draft, AI assistance was used
to improve sentence-level fluency and clarity. The author reviewed the
resulting edits and takes responsibility for the final text.

\bibliography{references}

\appendix

\section*{Appendix Roadmap}

The appendices are grouped into five parts. Parts~I--II document how the data
and interventions were made, including the verbatim instruments; Part~III
walks a single query end to end and then illustrates each headline claim with
a real archived case; Parts~IV--V give the statistical and reproducibility
record.

\begin{table*}[t]
\centering
\small
\begin{tabular}{@{}llp{0.62\textwidth}@{}}
\toprule
Part & App. & Contents \\
\midrule
I.\ \ The data pipeline & A--C & query set and acquisition; blinded evidence audit and pair construction; treatment generation and integrity gates \\
II.\ \ Instruments & D & all eight prompts verbatim, with archived hashes \\
III.\ \ Worked examples & E--F & one query traced end to end; then one real archived case per headline claim \\
IV.\ \ Statistical detail & G--K & estimators and inference; multiplicity; repeatability and generation noise; rank geometry; cross-model replication and missingness \\
V.\ \ Provenance & L--M & models, decoding, model-visible interface; archives, artifact, protocol amendments \\
\bottomrule
\end{tabular}
\caption{Appendix roadmap. Every claim in the body carries a breadcrumb to
the appendix that documents it.}
\label{tab:roadmap}
\end{table*}

\bigskip
\noindent\textbf{\large Part I \quad The Data Pipeline}

\section{Query Set and Acquisition}
\label{app:acquisition}

The acquisition frame samples 130 records from a manually revised 500-query
set, balanced 13 per topic across ten everyday domains and 26 per phrasing
family across five ordinary ways of asking (neutral fact bundles,
first-person scenarios, primary-source requests, plain-English questions,
terse search-style input). Table~\ref{tab:funnel} gives the resulting funnel.

\begin{table}[h]
\centering
\footnotesize
\begin{tabular}{@{}p{0.68\columnwidth}r@{}}
\toprule
Quantity & Value \\
\midrule
Queries selected / acquired / evidence-processed & 130 / 129 / 128 \\
Topics $\times$ phrasing forms & 10 $\times$ 5 \\
Final search calls / transport attempts & 346 / 350 \\
Unique documents / evidence-reviewed & 1{,}490 / 1{,}465 \\
Answer-bearing documents & 1{,}115 (76.1\%) \\
Candidate competing pairs (queries) & $\geq$1{,}894 (125) \\
Frozen cohort / replayed pairs & 114 / 113 \\
Answer-target families & 89 \\
GPT replay cells (planned / complete) & 452 / 452 \\
\bottomrule
\end{tabular}
\caption{Data funnel for the scaled wave. One query failed acquisition and
one acquired record failed evidence processing, both before pair selection.
The candidate-pair count is a lower bound because extraction caps pairs at 30
per query (13 queries saturated the cap). Of the 114 frozen pairs (nine to 13
per topic), the one treatment-infeasible pair leaves the replayed cohort with
eight pairs in its topic (legal/civic) and nine to 13 elsewhere.}
\label{tab:funnel}
\end{table}

First-exposure citation rates underlying Figure~\ref{fig:gradient}: 85.1\%
(rank 1, $n{=}276$ documents), 66.0\% (294), 54.0\% (300), 49.8\% (309),
42.8\% (311). These are descriptive; rank is confounded with relevance,
quality, and the agent's query choices.

\section{Evidence Audit and Pair Construction}
\label{app:pairs}

\paragraph{Answer-unit construction.} Each record of the manually revised
query set carries two human-authored fields: \texttt{required\_answer\_units}
(a short list of the facts an acceptable answer must cover) and an
\texttt{anchor\_fact\_bundle} summarizing the factual core. Before any
document is retrieved, a Grok-4.3 pass converts these fields into two to
eight atomic propositions per record (median five over the 130 records),
under instructions that forbid adding facts, thresholds, entities, or
recommendations not already expressed in the supplied fields; broad
categories become support criteria rather than invented specifics. The
resulting units are frozen to an append-only table whose SHA-256 enters the
run manifest, so units cannot drift after documents are seen. The units are
therefore human-sourced in content and model-normalized in form; the human
audit packet (Appendix~\ref{app:repro}) covers the downstream support
labels built on them.

\paragraph{Evidence audit.} A blinded evaluator (prompt~D.3) judges every
model-visible document text against its query's frozen units and must return
a verbatim quote, verified by exact string match against the archived text;
absent or merely paraphrased quotes are conservatively downgraded to
\texttt{unclear} and cannot make a pair eligible.

A candidate pair must satisfy all of: (i) both documents in the same
model-issued search call; (ii) verified answer-bearing support for a shared
unit in both; (iii) no material contradiction of that unit; (iv) no
duplicate or syndicated content; (v) metadata complete enough for integrity
diffs; (vi) answer-unit coverage differing by at most two units; (vii)
returned-text length ratio at least 0.25; (viii) text suitable for a faithful
prose/structure contrast.

\paragraph{One pair per query.} When several candidate pairs survive
screening for a query, the frozen pair is the one maximizing a deterministic
priority score ($5\times$shared-unit count $+\,3\times$content-length ratio
$-$ supported-unit-count difference), with ties broken by pair ID; selection
is topic-balanced under a per-topic cap, and manually rejected pairs are
replaced by the next-best candidate for the same query. The treatment target
within each pair is fixed by an outcome-blind hash of the seed, pair ID, and
sorted source IDs.

\paragraph{Human validation of shared-evidence labels.} A human author
adjudicated, blind to document identities, ranks, and outcomes, one
hash-chosen decisive shared unit for every analyzed pair (113 items, both
documents' verified quotes) plus a 14-item borderline supplement drawn from
the machine evaluator's own hard cases (partial-completeness, indirect, or
low-confidence labels). Of 254 document-level judgements, 94.9\% were
supports or partial; at the pair level, 103 of 113 pairs (91.2\%) were
confirmed as genuine shared-evidence competitions and ten were invalidated,
typically because the two quotes were topically related without asserting
the same proposition. Invalid pairs were distributed proportionally across
effect strata (7 of 83 no-change pairs, 3 of 30 discordant pairs).
Excluding them strengthens every headline estimate: structure incidence
$+5.9\pp$ (bootstrap 95\% CI $[+0.3,+12.0]$; sign-flip $p{=}.066$; 81
families), structure citation count $+0.51$ ($[+0.17,+0.88]$; $p{=}.005$),
and rank incidence $+9.6\pp$ ($[+2.8,+16.7]$; $p{=}.011$). These
audit-restricted estimates are sensitivity analyses; the declared primary
results in Section~\ref{sec:results} are unchanged. Adjudications, the
audit interface, and per-item decisions ship in the artifact.

\section{Treatment Generation and Integrity}
\label{app:integrity}

Both renderings are generated in a single call from the same frozen source
text (prompt~D.4), so neither arm can be tuned against the other. The
generator is the Grok-4.3 deployment (temperature $0$, 20{,}000-token output
cap, one protocol retry), with a GPT-5.4 fallback after repeated protocol
failures. Exactly one accepted treatment came from the fallback (pair
\texttt{P2A6EA681D8AD}); excluding it leaves the primary estimate unchanged
($+4.55\pp$ over 88 families). The single pair where both generators failed
is the declared pre-outcome attrition. Deterministic validators then
enforce: at least 120 words per arm;
word counts within 25\%; no citation handles inside renderings; zero
structural markers in the prose arm; at least two headings and two list or
table rows in the structured arm. A separate blinded model pass
(prompt~D.5) must pass all seven fidelity dimensions.

At replay time the frozen target text hash is re-verified, and every trial
stores pre- and post-intervention hashes for the transcript, tool results,
each source object, and the opaque-handle map. The scaled GPT run records
zero live search calls, zero citation repairs, zero missing cells, and zero
integrity failures; 31 failed transport attempts across 25 pairs were retried
under unchanged frozen trial definitions.

\bigskip
\noindent\textbf{\large Part II \quad Instruments}

\section{Prompt Catalogue}
\label{app:prompts}

Eight prompts define the study. Each is reproduced below \emph{complete and
verbatim} (LaTeX-escaped only); the header of each card shows the leading
eight hex digits of the file's SHA-256, which is recorded in every run
manifest and shipped in the artifact.

\paragraph{D.1 Acquisition agent (system).}
Defines the agent contract during live acquisition: search before answering, bounded turns and calls, strict-JSON actions, opaque citation handles.

\begin{promptcard}{D.1 \quad agent\_loop\_system}{8ef78b13}
You are the tool-using answer agent inside a production AI search product.\\
\smallskip
You answer user questions with web evidence. You have exactly one tool:\\
\smallskip
search\_web(query)\\
\smallskip
The search\_web tool returns web search results with source IDs. You must call\\
search\_web at least once before you answer. You may call multiple searches in\\
one turn when the question has distinct aspects.\\
\smallskip
Return strict JSON only. Do not write prose outside JSON.\\
\smallskip
Allowed actions:\\
\smallskip
1. Search:\\
\{\\
  ''action'': ''search\_web'',\\
  ''tool\_calls'': [\\
    \{\\
      ''query'': ''short web search query'',\\
      ''reason'': ''why this search is needed''\\
    \}\\
  ]\\
\}\\
\smallskip
2. Respond:\\
\{\\
  ''action'': ''respond'',\\
  ''answer'': ''final answer with [[cite:SOURCE\_ID]] citations'',\\
  ''confidence'': ''low|\allowbreak{}medium|\allowbreak{}high'',\\
  ''used\_source\_ids'': [''S01'']\\
\}\\
\smallskip
Rules:\\
- First turn must call search\_web.\\
- Use at most 3 search\_web calls per turn.\\
- Use only source IDs returned by search\_web.\\
- Cite factual claims with [[cite:SOURCE\_ID]].\\
- If search results are insufficient, call search\_web again with better queries.\\
- Do not repeat searches that are substantially the same as earlier searches.\\
- If available sources are relevant but imperfect, answer with calibrated caveats\\
  instead of searching indefinitely.\\
- On the final allowed turn, respond using the best available sources unless no\\
  retrieved source is relevant to the user query.\\
- Keep answers concise and grounded.\\
\end{promptcard}

\paragraph{D.2 Answer generation (system).}
Used for every replayed trial. Deliberately generic product-style instruction: the model is never told documents are being compared.

\begin{promptcard}{D.2 \quad answer\_generation\_system}{bfb73bf1}
You are the answering component of a production AI search product, similar to\\
ChatGPT Search, Google AI Mode, or Perplexity.\\
\smallskip
Answer the user's question using only the provided source snapshots. Cite every\\
factual sentence with source IDs in the exact format [[cite:SOURCE\_ID]].\\
\smallskip
Do not cite sources that are not provided. Do not cite search snippets unless\\
their page content was extracted and provided as a source snapshot. If the\\
sources are insufficient, say what is missing instead of guessing.\\
\end{promptcard}

\paragraph{D.3 Evidence audit (system).}
Establishes the answer-bearing labels that define the study population; note the explicit blinding and anti-position instruction.

\begin{promptcard}{D.3 \quad document\_support\_matrix\_system}{8c88faa3}
You are a blinded evidence-support auditor for a citation-selection study.\\
\smallskip
For each neutral document and every frozen answer unit, decide whether the\\
supplied model-visible passages establish that unit. Judge only the supplied\\
passages. Do not browse, use outside knowledge, infer omitted page content, or\\
reward a document based on order. Titles, domains, URLs, citation outcomes, and\\
publisher identities are intentionally hidden.\\
\smallskip
Support labels:\\
- supports: the passages directly justify the complete unit, including its\\
  material qualifications;\\
- partial: they justify only part of the unit or omit a material qualification;\\
- does\_not\_support: they are related but do not establish the unit;\\
- contradicts: they materially conflict with the unit;\\
- unclear: the supplied passages are insufficient to decide.\\
\smallskip
For `supports` or `partial`, return at least one supplied passage\_id and a short\\
verbatim evidence\_quote copied from those passages. Never fabricate a passage\\
ID or quotation. Use `complete`, `partial`, `not\_applicable`, or `unclear` for\\
completeness and `direct`, `indirect`, `none`, or `unclear` for directness.\\
\smallskip
Return strict JSON only:\\
\{\\
  ''documents'': [\\
    \{\\
      ''document\_id'': ''neutral ID'',\\
      ''units'': [\\
        \{\\
          ''unit\_id'': ''U01'',\\
          ''support'': ''supports|\allowbreak{}partial|\allowbreak{}does\_not\_support|\allowbreak{}contradicts|\allowbreak{}unclear'',\\
          ''passage\_ids'': [''passage ID''],\\
          ''evidence\_quote'': ''verbatim supplied text or empty string'',\\
          ''completeness'': ''complete|\allowbreak{}partial|\allowbreak{}not\_applicable|\allowbreak{}unclear'',\\
          ''directness'': ''direct|\allowbreak{}indirect|\allowbreak{}none|\allowbreak{}unclear'',\\
          ''contradiction'': ''none|\allowbreak{}minor|\allowbreak{}major|\allowbreak{}unclear'',\\
          ''confidence'': ''low|\allowbreak{}medium|\allowbreak{}high''\\
        \}\\
      ]\\
    \}\\
  ]\\
\}\\
\end{promptcard}

\paragraph{D.4 Treatment generation (system).}
Produces both arms jointly; the editor is forbidden from reasoning about citations.

\begin{promptcard}{D.4 \quad structure\_rewrite\_system}{23dbbdb8}
You are the treatment-material editor for a controlled study of how document\\
structure affects source citation. You are not an answer engine and must not\\
judge which source should be cited.\\
\smallskip
Create two editorial renderings of the supplied frozen search-result text:\\
\smallskip
1. `polished\_prose`: clear continuous prose in ordinary paragraphs. Do not use\\
   headings, bullets, numbered lists, tables, labels, or callout blocks.\\
2. `polished\_structured`: equally polished content organized with descriptive\\
   Markdown headings, short coherent paragraphs, and bullets or a table only\\
   where they naturally improve scanning.\\
\smallskip
The two renderings must be semantically equivalent. Preserve the same factual\\
claims, quantities, named entities, qualifications, uncertainty, exceptions,\\
procedures, warnings, and source attribution. Do not add facts from memory,\\
infer missing details, resolve ambiguity, or tailor one rendering more closely\\
to the user query. Remove navigation fragments and duplicated boilerplate only\\
if both renderings remove the same material. Keep the two versions within 25\%\\
of each other in word count and information density.\\
\smallskip
Use unnumbered bullets in the structured version unless sequence is itself a\\
fact. Never include citation handles such as `[[cite:...]]`.\\
\smallskip
Return one strict JSON object only:\\
\{\\
  ''polished\_prose'': \{\\
    ''text'': ''complete prose rendering'',\\
    ''word\_count'': 0\\
  \},\\
  ''polished\_structured'': \{\\
    ''text'': ''complete structured rendering'',\\
    ''word\_count'': 0\\
  \},\\
  ''claim\_inventory'': [\\
    \{\\
      ''claim'': ''factual claim present in both renderings'',\\
      ''material\_qualifiers'': [''preserved caveat or qualification'']\\
    \}\\
  ],\\
  ''editor\_notes'': [''brief note about symmetric boilerplate removal, if any'']\\
\}\\
\end{promptcard}

\paragraph{D.5 Fidelity audit (system).}
A separate blinded auditor pass gates every treatment on seven dimensions; one failure rejects the pair.

\begin{promptcard}{D.5 \quad structure\_fidelity\_audit\_system}{b0a455fa}
You are an independent fidelity auditor for controlled document-structure\\
treatments. You do not improve the rewrites and do not predict citations.\\
\smallskip
Compare each polished-prose and polished-structured pair against the frozen\\
source text and against each other. A variant is eligible only when all of the\\
following pass:\\
\smallskip
- `claim\_equivalence`: the two renderings express the same factual claims and\\
  answer-bearing coverage;\\
- `no\_new\_facts`: neither rendering introduces unsupported factual content;\\
- `caveat\_preservation`: material uncertainty, exceptions, warnings, and\\
  procedural conditions are equivalent;\\
- `quantitative\_fidelity`: quantities, dates, thresholds, units, and named\\
  entities agree with the source and each other;\\
- `attribution\_fidelity`: source identity and attribution present in the source\\
  are retained equivalently;\\
- `structure\_contrast`: prose is genuinely continuous prose and structured text\\
  has useful editorial organization beyond cosmetic bullet conversion;\\
- `length\_balance`: information density and length are comparable, with no arm\\
  receiving a materially richer summary.\\
\smallskip
Return strict JSON only:\\
\{\\
  ''reviews'': [\\
    \{\\
      ''variant\_id'': ''provided ID'',\\
      ''claim\_equivalence'': ''pass|\allowbreak{}fail'',\\
      ''no\_new\_facts'': ''pass|\allowbreak{}fail'',\\
      ''caveat\_preservation'': ''pass|\allowbreak{}fail'',\\
      ''quantitative\_fidelity'': ''pass|\allowbreak{}fail'',\\
      ''attribution\_fidelity'': ''pass|\allowbreak{}fail'',\\
      ''structure\_contrast'': ''pass|\allowbreak{}fail'',\\
      ''length\_balance'': ''pass|\allowbreak{}fail'',\\
      ''eligible'': true,\\
      ''issues'': [''specific issue, empty when none''],\\
      ''rationale'': ''concise evidence-grounded audit''\\
    \}\\
  ]\\
\}\\
\end{promptcard}

\paragraph{D.6 Shared-unit citation alignment (system).}
Added in revision (Section~\ref{sec:allocation}). Sees only the query, units, and claim sentences; never arm, rank, or document identity.

\begin{promptcard}{D.6 \quad shared\_unit\_alignment\_system}{abd01507}
You are a meticulous citation-alignment auditor for a research study.\\
\smallskip
You will receive:\\
1. a user query;\\
2. a list of SHARED ANSWER UNITS: atomic facts that two candidate documents\\
   were both independently verified to support;\\
3. a list of CLAIM EXCERPTS from a generated answer. Each excerpt is the\\
   sentence (or short passage) to which a citation of one specific document\\
   was attached.\\
\smallskip
For each claim excerpt, decide whether the fact the excerpt asserts\\
corresponds to one of the shared answer units.\\
\smallskip
Rules:\\
- Return ''U01''-style unit IDs only when the excerpt's asserted fact is the\\
  same fact as the unit, allowing paraphrase but not topic drift.\\
- Return ''none'' when the excerpt asserts a fact that is not any shared unit\\
  (for example a document-specific detail, a different subtopic, or generic\\
  framing).\\
- Return ''unclear'' only when the excerpt is too fragmentary to judge.\\
- Judge each excerpt independently. Do not reward or penalize the answer.\\
\smallskip
Respond with ONE strict JSON object and nothing else:\\
\{''alignments'': [\{''citation\_index'': <int>, ''matched\_unit\_id'': ''U01'' |\allowbreak{} ''none'' |\allowbreak{} ''unclear'', ''reason'': ''<one short sentence>''\}]\}\\
Include exactly one entry for every provided citation\_index.\\
\end{promptcard}

\paragraph{D.7 Unit expression and allocation (system).}
Added in revision. Judges, per shared unit, whether the answer expresses it and which neutral source tag is credited.

\begin{promptcard}{D.7 \quad unit\_allocation\_system}{d98d2bc1}
You are a meticulous evidence-expression auditor for a research study.\\
\smallskip
You will receive:\\
1. a user query;\\
2. a list of SHARED ANSWER UNITS: atomic facts that two candidate documents\\
   were both independently verified to support;\\
3. a generated ANSWER in which every citation marker has been replaced by a\\
   neutral source tag: [S1], [S2] (the two candidate documents, in a hidden\\
   random order), or [OTHER] (any other source).\\
\smallskip
For each shared answer unit, decide:\\
- ''expressed'': does any sentence of the answer assert this fact?\\
  ''yes'' (asserted, allowing paraphrase), ''no'' (absent), or ''partial''\\
  (only a fragment or weakened form appears).\\
- ''cited\_sources'': which source tags are attached to the sentence(s) that\\
  express this unit. List any of ''S1'', ''S2'', ''OTHER''; use an empty list when\\
  the expressing sentences carry no tag or the unit is not expressed.\\
\smallskip
Rules:\\
- Judge expression from the answer text alone; do not use outside knowledge.\\
- Attribute a tag to a unit only when the tag is attached to a sentence that\\
  expresses that unit, not merely nearby.\\
- Judge each unit independently. Do not reward or penalize the answer.\\
\smallskip
Respond with ONE strict JSON object and nothing else:\\
\{''units'': [\{''unit\_id'': ''U01'', ''expressed'': ''yes'' |\allowbreak{} ''no'' |\allowbreak{} ''partial'', ''cited\_sources'': [''S1'']\}]\}\\
Include exactly one entry for every provided unit\_id.\\
\end{promptcard}

\paragraph{D.8 Citation repair (user).}
Fires once when a final answer has no valid citation handle; restates the protocol without re-opening search. Never triggered in the scaled GPT runs; Grok needed it for 121 of 219 answers.

\begin{promptcard}{D.8 \quad factorial\_replay\_citation\_repair\_user}{becb36ee}
The answer draft violated the citation protocol:\\
\smallskip
\{\{ violation \}\}\\
\smallskip
Return a corrected final answer to the original user query. Preserve the\\
substance of the draft, cite factual claims using only the opaque source handles\\
that appeared in the archived search tool results, and use exactly\\
`[[cite:SOURCE\_ID]]` for each citation. Do not request another search.\\
\end{promptcard}

\bigskip
\noindent\textbf{\large Part III \quad Worked Examples}

\section{End-to-End Walkthrough: One Query}
\label{app:walkthrough}

This section follows a single record, \texttt{CCV2\_TRV06\_02\_first\_person\_scenario},
from live acquisition to final outcome. It is the same pair used for the rank
illustration (Appendix~\ref{app:example-rank}) and the ablation excerpt
(Appendix~\ref{app:example-ablation}).

\paragraph{Stage 1: the query.} Drawn from the travel topic in the
first-person-scenario phrasing family: \emph{``I am returning to the U.S.\
from abroad. Should I use Mobile Passport Control or Global Entry if I do not
travel often?''}

\paragraph{Stage 2: the agent searches.} Given only the D.1 contract, the
GPT-5.4 agent issued three parallel searches in its first turn and then
answered:

\begin{quote}\footnotesize
1. \texttt{CBP Mobile Passport Control official Global Entry official fees
interview current 2025 infrequent traveler comparison}\\
2. \texttt{official CBP Mobile Passport Control eligibility participating
airports no pre-approval free app official}\\
3. \texttt{official CBP Global Entry fee interview required benefits TSA
PreCheck official current}
\end{quote}

\paragraph{Stage 3: the returned call.} The first call returned five results.
Positions 1 and 5 became the competing pair. Both are published by
\texttt{cbp.gov}, which is what makes this pair such a clean rank test:

\begin{quote}\footnotesize
\textbf{1.} \emph{Mobile Passport Control (MPC)} \hfill \texttt{R074BA2}\\
2.\ 3.\ (other \texttt{cbp.gov} pages)\\
4.\ (\texttt{ttp.cbp.dhs.gov})\\
\textbf{5.} \emph{Global Entry Frequently Asked Questions} \hfill \texttt{R2A8AAE}
\end{quote}

\paragraph{Stage 4: blinded evidence audit.} Judging blinded text only, the
evaluator found both documents independently answer-bearing on the same
units, each with a verified verbatim quote. The units vary in how tightly
they pin down a single proposition, and the table is ordered accordingly:

\begin{table}[h]
\centering
\footnotesize
\begin{tabular}{@{}lp{0.30\columnwidth}p{0.30\columnwidth}@{}}
\toprule
Unit & \texttt{R074BA2} (MPC) & \texttt{R2A8AAE} (Global Entry) \\
\midrule
U02 cost & ``Cost \textbar{} Free \textbar{} \$120 for 5-year
membership'' & ``pay the \$120 Global Entry application fee'' \\
U06 app use & ``Must download and use the MPC app'' & ``The Global Entry
Mobile App will allow members to validate their arrival\ldots'' \\
U01 eligibility & ``Available to U.S.\ citizens, U.S.\ lawful permanent
residents\ldots'' & ``Applicants may not qualify for Global Entry
participation if they:'' \\
\bottomrule
\end{tabular}
\caption{Verified support quotes for the three shared answer units, ordered
from tightest to loosest. The cost unit (both documents state the \$120 fee
outright) is the one the human adjudicator treated as decisive; the
eligibility unit is a broader comparative criterion, where the two quotes
address the same question without asserting one identical sentence.}
\label{tab:walkthrough-evidence}
\end{table}

\noindent This gradient is precisely what the pair-level human audit
adjudicates. The adjudicator saw the cost unit as this pair's decisive item
and scored both documents \emph{supports} with genuine competition, so the
pair is one of the 103 confirmed competitions; ten other pairs failed that
test and are excluded in the sensitivity analysis of
Appendix~\ref{app:pairs}.

\paragraph{Stage 5: freezing and target assignment.} The pair passed all
eight screening criteria and was frozen as \texttt{PC812B36BC572}, the single
pair representing this query. The outcome-blind hash designated the Global
Entry FAQ (originally rank~5) as the target; the MPC page is the never-edited
competitor.

\paragraph{Stage 6: treatments.} Both renderings were generated jointly from
the target's frozen text and passed all seven audit dimensions.
Appendix~\ref{app:example-ablation} shows the prose arm against the
word-preserving mechanical list arm used in the ablation.

\paragraph{Stage 7: replay.} Four trials crossed target rendering with rank
assignment over the byte-identical transcript. The result is
Table~\ref{tab:example-rank}: the target was cited when promoted to rank~1 in
\emph{both} text arms and cited in neither arm when left at rank~5, while its
sibling page drew 5--7 citations in every condition.

\paragraph{Stage 8: alignment.} The citation the promoted target earned was
attached to the sentence ``\ldots You want the added benefit of \textbf{TSA
PreCheck eligibility} bundled with it,'' which the blinded alignment auditor
scored against this pair's shared units.

\section{Additional Worked Examples}
\label{app:example}

Each example below is a real archived pair chosen to illustrate one headline
claim, and every one was confirmed as a genuine competition by the blinded
human adjudication. Excerpts are abbreviated for space; the model saw the
full texts.

\subsection{Rank: same publisher, different slot}
\label{app:example-rank}

Table~\ref{tab:example-rank} is the cleanest available rank illustration
because both competing documents come from the \emph{same publisher}
(\texttt{cbp.gov}), so source authority, domain reputation, and house style
are held constant by construction. The documents differ only in which slot
they occupy. The target originally sat at rank~5 against a sibling page at
rank~1: a full-window, four-slot swap, the largest displacement stratum in
Table~\ref{tab:displacement}.

\begin{table}[h]
\centering
\small
\begin{tabular}{@{}lcc@{}}
\toprule
Target citations & Target at rank 1 & Target at rank 5 \\
\midrule
Polished prose      & 1 & 0 \\
Polished structured & 1 & 0 \\
\bottomrule
\end{tabular}
\caption{Pair \texttt{PC812B36BC572} (walkthrough of
Appendix~\ref{app:walkthrough}). Both documents are \texttt{cbp.gov} pages
returned by the same Exa call. The competitor drew 5--7 citations regardless
of condition.}
\label{tab:example-rank}
\end{table}

\subsection{Structure: same facts, different organization}
\label{app:example-structure}

This is the clearest structure contrast in the cohort, and it doubles as an
illustration of the shared-evidence construct at its strictest. Both
documents are \texttt{travel.state.gov} pages returned by the same Exa call,
so publisher authority and house style are held constant, and their verified
quotes for the shared unit \emph{Passport validity periods are
destination-specific} are near-paraphrases of one another:

\begin{quote}\footnotesize
\emph{Competitor} (\texttt{R61B1FF}, passport FAQ). ``Some destinations
require that your passport be valid at least 6 months beyond the dates of
your trip.''

\emph{Target} (\texttt{RA2FA82}, after-you-apply page). ``Some foreign
destinations require that your passport have at least 6 months of validity
beyond the dates of your trip.''
\end{quote}

\noindent The blinded human adjudicator scored both as \emph{supports} with
genuine competition. Either page could be cited for the six-month rule, so
which one the engine credits is pure allocation. Table~\ref{tab:example}
shows what the rendering did to that choice: the target was cited in
\emph{both} structured cells and \emph{neither} prose cell, while the
competitor was cited in all four cells. The competitor never leaves the
answer; the target joins it only when its text is structured.

\begin{table*}[h]
\centering
\small
\begin{tabular}{@{}p{0.46\textwidth}p{0.50\textwidth}@{}}
\toprule
\multicolumn{2}{@{}p{0.97\textwidth}@{}}{\textbf{Query:} ``Explain passport
validity and blank-page requirements for international travel in plain
English\ldots'' Target: the after-you-apply page at rank 3; competitor: the
passport FAQ at rank 2 (same Exa call, adjacent slots). Target citations: 3
and 4 markers in the two structured cells, 0 in both prose cells; competitor
cited in all four (1--7 markers). Pair \texttt{P2B841667E0D2}.} \\
\midrule
\emph{Polished prose (excerpt)} & \emph{Polished structured (excerpt)} \\
``After receiving a new passport travelers should understand its validity
periods and preparation steps for international travel. Some destinations
require the passport to remain valid for at least six months beyond the trip
dates and certain airlines will deny boarding if this condition is unmet.
Passports issued to individuals age sixteen or older are valid for ten years
while those issued to children under sixteen are valid for five
years\ldots'' &
``\textbf{\#\# Passport Validity for Travel}\newline Some destinations
require that a passport remain valid for at least six months beyond the dates
of the trip. Airlines may refuse boarding if this requirement is not
met.\newline
\texttt{-} Passports issued to those age sixteen or older are valid for ten
years.\newline
\texttt{-} Passports issued to those under age sixteen are valid for five
years\ldots'' \\
\addlinespace
\multicolumn{2}{@{}p{0.97\textwidth}@{}}{\emph{Answer sentence that cited the
structured target:} ``\ldots U.S.\ State Department guidance says some foreign
destinations require at least 6 months of extra validity, and airlines may
refuse boarding if you do not meet that rule.''} \\
\bottomrule
\end{tabular}
\caption{A real archived pair, two pages from the same publisher. Both
renderings carry the same facts in the same order; only the structured one
drew citations, in both rank conditions.}
\label{tab:example}
\end{table*}

\subsection{The mechanical ablation, verbatim}
\label{app:example-ablation}

The ablation's two arms share an identical word sequence; only line breaks
and list markers differ, so the model-visible character stream (though not
the lexical content) does change. From the same Global Entry target as
Appendix~\ref{app:walkthrough}:

\begin{quote}\footnotesize
\emph{Prose arm.} ``Global Entry is a risk-based approach to facilitate the
entry of pre-approved travelers. Applicants may not qualify for Global Entry
participation if they provide false or incomplete information on the
application, have been convicted of any criminal offense\ldots''

\emph{Mechanical list arm.} ``\texttt{-} Global Entry is a risk-based
approach to facilitate the entry of pre-approved travelers.\newline
\texttt{-} Applicants may not qualify for Global Entry participation if they
provide false or incomplete information on the application, have been
convicted of any criminal offense\ldots''
\end{quote}

No word is added, removed, or reordered; a programmatic check enforces
word-sequence identity for all 113 targets. This is the entire manipulation that produced
the $+6.7\pp$ ablation estimate in Section~\ref{sec:structure}.

\subsection{Citation alignment: shared versus target-specific}
\label{app:example-alignment}

The alignment audit's distinction is concrete. For pair
\texttt{P50955652FFFC} (``when might acetaminophen be a better choice than
ibuprofen?''), the pair's two verified shared units concern ibuprofen's
stomach cautions (U04) and drug-interaction caveats (U05). One structured-arm
answer cited the target four times, which the blinded auditor split evenly:

\begin{quote}\footnotesize
\emph{Aligned to U04.} ``NSAIDs such as ibuprofen can cause \textbf{stomach
ulcers and bleeding}, and the risk is higher in older adults, people with
prior ulcers\ldots''

\emph{Not a shared unit.} ``\textbf{Do not exceed the labeled dose.} Too much
acetaminophen can cause \textbf{severe liver damage}\ldots''
\end{quote}

The first citation credits the target for evidence its competitor also
carried. This is the contested allocation this paper studies. The second
credits it for target-specific content no competitor offered, which the aggregate
citation-count outcome would otherwise conflate. Across the corpus this split
is 48.6\% versus 50.7\% (0.7\% unclear), and the structure effect survives
restriction to the aligned portion (Section~\ref{sec:allocation}).

\bigskip
\noindent\textbf{\large Part IV \quad Statistical Detail}

\section{Estimators and Inference}
\label{app:estimators}

For pair $i$ with text arm $c\in\{P,S\}$ and assigned rank $r\in\{H,L\}$, all
four cells are observed, so each contrast is a within-pair difference of cell
means; no modeling assumptions link pairs. Pair effects are averaged within
answer-target family $g$, giving $n{=}89$ independent units from 113 pairs.

Confidence intervals resample families with replacement (20{,}000
replicates, percentile method). Two-sided $p$-values come from a sign-flip
permutation test on the nonzero family effects: with $m \leq 20$ discordant
families we enumerate all $2^m$ sign assignments exactly via Gray-code
updates; beyond that we draw 200{,}000 seeded Monte Carlo flips and report
$(\text{extreme}+1)/(\text{reps}+1)$. All seeds are archived.

Because one generation is drawn per cell, decoding noise is absorbed into the
family-level variance the bootstrap estimates: inference remains valid for
the expected effect but is conservative in power. Appendix~\ref{app:repeat}
quantifies that noise directly.

\section{Secondary Outcomes and Multiplicity}
\label{app:secondary}

\begin{table}[h]
\centering
\small
\begin{tabular}{@{}lrr@{}}
\toprule
Secondary test (scaled wave) & Raw $p$ & Holm $p$ \\
\midrule
Citation count: structure effect & .0022 & \textbf{.033} \\
Pair-citation share: rank, structured arm & .0001 & \textbf{.0018} \\
First pair citation: structure effect & .030 & .332 \\
Incidence: rank main effect & .035 & .350 \\
Incidence: rank effect, prose arm & .041 & .365 \\
Citation count: rank effect, prose arm & .0045 & .063 \\
First pair citation: rank, prose arm & .0059 & .076 \\
Pair-citation share: rank, prose arm & .0074 & .089 \\
\bottomrule
\end{tabular}
\caption{The eight smallest raw $p$-values among the 16 declared secondary
tests (all others have Holm $p \geq .33$). Bold: survives Holm at $.05$. The
surviving effects are the target citation-count increase under structure
($+0.50$, 95\% CI $[+0.20,+0.84]$) and the rank effect on within-pair citation
share under structured rendering ($+10.5\pp$, defined over the 72 families with at
least one within-pair citation in both rank cells).}
\label{tab:holm}
\end{table}

Answer-level diagnostics (structured minus prose, clustered): total citation
markers $-0.49$ (95\% CI $[-1.6,+0.6]$); unique cited sources $-0.05$
($[-0.20,+0.11]$); answer characters $-42$ ($[-140,+53]$). None differs
reliably from zero, supporting the reallocation interpretation in
Section~\ref{sec:allocation}.

\paragraph{Count and share baselines.} Behind the headline contrasts:
target citation-count cell means are prose 2.69 / structured 3.12 per
answer (competitor: 4.01 under both arms); the higher-ranked target's mean
within-pair citation share is 48.4\% vs.\ 37.9\% (higher vs.\ lower)
under structured rendering over the 72-family share population, and 40.1\% vs.\
32.4\% under prose (74 families). The share population conditions on
realized within-pair citations in both rank cells and is read as
descriptive (Section~\ref{sec:allocation}).

\paragraph{Leave-one-topic-out (named).} Excluding each topic in turn, the
primary estimate is: consumer electronics $+5.9$; cooking/food safety
$+4.4$; education/study methods $+1.6$; low-acuity health $+4.7$; home
DIY $+4.4$; home products $+4.7$; legal/civic $+5.2$; personal finance
$+6.3$; software $+4.0$; travel $+3.8\pp$. The education topic contributes
most: its families concentrate positive effects, and removing them drops
the estimate by about $2.9\pp$, within the deletion-check range reported
in Section~\ref{sec:structure} but worth naming.

\paragraph{Answer-target family construction.} Families equal the
\texttt{answer\_target\_id} assigned when queries were authored (before
any document, transcript, or outcome existed), so membership is
deterministic and outcome-blind. Of the 89 families, 65 contain one
analyzed pair and 24 contain two; no family contains more.

\paragraph{What the count effect is made of (post hoc).} To rule out the
degenerate reading of $+0.50$ as a repeated marker on a single sentence, we
recomputed the effect over structural variants of the outcome (89 clusters,
20{,}000-replicate bootstrap, sign-flip $p$; all post hoc): \emph{distinct
answer sentences} citing the target rise by $+0.50$, from 2.69 to 3.11 per
answer ($[+0.20,+0.83]$, $p{=}.002$); \emph{unique shared units} credited
to the target (from the blinded alignment audit, zero for uncited answers)
rise by $+0.23$, from 1.24 to 1.47 ($[+0.09,+0.37]$, $p{=}.002$); and among
the 44 pairs whose target is cited in all four cells, which removes the
admission margin entirely, the count effect is $+0.60$ ($[0.00,+1.25]$,
$p{=}.075$, 39 clusters). The family-level distribution of count effects is
37 positive, 29 zero, and 23 negative (median $0.00$, quartiles
$[-0.25,+1.00]$, range $[-3.0,+6.5]$): a broad shift, not a few outlier
families (artifact file \texttt{credit\_supplements.json}).

\section{Repeatability and Generation Noise}
\label{app:repeat}

\begin{table}[h]
\centering
\small
\begin{tabular}{@{}lr@{}}
\toprule
Agreement between generations (120 cells) & Rate \\
\midrule
Target cited (binary) & 85.0\% \\
First pair citation & 87.5\% \\
Exact cited-pair-member set & 78.3\% \\
Exact target citation count & 61.7\% \\
\bottomrule
\end{tabular}
\caption{Cell-level agreement between the primary generation and an
independent decoding pass for 30 hash-selected answer-target families.
Transitions: 6 cells gained a target citation, 12 lost one, 102 unchanged.}
\label{tab:repeat}
\end{table}

The selected families' first-generation structure effect is $+6.7\pp$; the
regeneration gives $+3.3\pp$ (95\% CI $[-5.0,+11.7]$), and the two-generation
average $+5.0\pp$ ($[-4.2,+13.3]$; $p{=}.359$). Exact pair-effect agreement:
17/30.

\paragraph{Ablation repeatability.} The word-preserving ablation received
the same treatment: fresh generations of all four cells for the identical
30 hash-chosen families (120 trials, run
\texttt{20260726-105808}; zero repairs or integrity failures; binary
agreement 90.0\%, exact counts 63.3\%, transitions symmetric 6/6). On this
subsample the ablation effect is $-1.7\pp$ in its original generation,
$-8.3\pp$ regenerated, $-5.0\pp$ averaged ($[-11.7,+0.8]$, $p{=}.22$).
This is opposite in sign to the full-cohort $+6.7\pp$ and to the
generated-rewrite contrast on the very same families
($+6.7$/$+3.3\pp$). The full-cohort ablation estimate is unchanged, but its
mechanism reading rests on a single-generation cohort estimate whose
subsamples are unstable, exactly as the variance decomposition below predicts.

\paragraph{Count-effect repeatability.} The citation-count effect, which is
the multiplicity-controlled secondary result, was recomputed on the same 30
families: $+0.54$ ($[+0.16,+0.96]$, $p{=}.014$) in the original generation,
$+0.55$ ($[-0.03,+1.22]$, $p{=}.114$) in the fresh generation, and $+0.55$
($[+0.16,+0.98]$, $p{=}.013$) averaging both. The aggregate magnitude
is nearly unchanged while family-level values correlate only $r{=}.26$
(direction agreement 63\%), the same aggregate-stable, family-noisy pattern
as incidence
(artifact file \texttt{round5\_supplements.json}).

\paragraph{Cross-wave family correlation.} Over the 49 answer-target
families recurring across waves (through different queries and transcripts),
incidence effects correlate $r{=}-.445$; a 100{,}000-draw permutation of the
pairing puts two-sided $p{=}.004$, so the anti-correlation is beyond chance.
It rests, however, on sparse discordance: only 8 pilot and 12 scale families
are nonzero at all, and just 4 are nonzero in both waves. The reading we
adopt is that family effects are properties of a specific frozen transcript,
not of the underlying fact. This interpretation is consistent with the
estimand's conditioning and with the 45\% decoding-noise share above.

\paragraph{Variance decomposition.}
Writing $e_{g}$ for a family's pair effect in generation $g$, the
method-of-moments estimate $\widehat{\sigma}^2_{\text{gen}} =
\overline{(e_1-e_2)^2}/2$ over the 30 twice-generated families gives
$\widehat{\sigma}^2_{\text{gen}} = .050$ against a single-draw pair-effect
variance of $.112$: decoding noise alone accounts for approximately 45\% of
the variance a one-generation design must absorb, with the remainder
reflecting between-family heterogeneity. Averaging $k$ generations per cell
would shrink the noise component by $1/k$, which is why the next confirmatory
design repeats every cell.

\section{Rank Geometry}
\label{app:geometry}

The 113 analyzed pairs occupy every absolute-position combination available
in a five-result call (higher slot--lower slot: count): 1--2: 17, 1--3: 11,
1--4: 12, 1--5: 10, 2--3: 14, 2--4: 8, 2--5: 12, 3--4: 13, 3--5: 12, 4--5: 4.
Mean swap displacement is 1.96 slots. By higher slot, the exploratory
clustered rank effect is $+16.0\pp$ when the higher position is rank~1 (50
pairs), $-4.4\pp$ at rank~2 (34), $+6.0\pp$ at rank~3 (25), and $0.0\pp$ at
rank~4 (4). Table~\ref{tab:displacement} gives the displacement-graded
breakdown; Table~\ref{tab:arms} gives the underlying cell rates.

\begin{table}[h]
\centering
\small
\begin{tabular}{@{}lrrl@{}}
\toprule
Swap distance & Pairs & Effect & 95\% CI \\
\midrule
1 slot  & 48 & $-3.3\pp$  & $[-12.2,+5.6]$ \\
2 slots & 31 & $+13.4\pp$ & $[0.0,+26.8]$ \\
3 slots & 24 & $+13.0\pp$ & $[+2.2,+26.1]$ \\
4 slots & 10 & $+30.0\pp$ & $[0.0,+60.0]$ \\
\bottomrule
\end{tabular}
\caption{Exploratory rank effect by swap distance (clustered over 45/28/23/10
answer-target families respectively, averaged over text arms). Cells are
small and none of these subgroup estimates is a corrected test.}
\label{tab:displacement}
\end{table}

\paragraph{Held-out confirmation geometry.} The 56 held-out pairs span
displacements of 1/2/3/4 slots with 21/15/11/9 pairs (mean 2.14, vs.\ 1.96
in the scaled wave), and their per-stratum effects are $0.0$, $-6.7$,
$+9.1$, and $0.0\pp$, respectively. They are near zero everywhere, including
the nine full-window pairs. The held-out zero is therefore not a composition
artifact; the operative differences from the scaled wave are text regime
(original vs.\ rewritten target text) and sampling width
(artifact file \texttt{round5\_supplements.json}).

\begin{table}[h]
\centering
\small
\begin{tabular}{@{}lcc@{}}
\toprule
Target incidence & Higher rank & Lower rank \\
\midrule
Polished prose & .584 & .487 \\
Polished structured & .602 & .558 \\
\bottomrule
\end{tabular}
\caption{Scaled-wave cell rates: share of trials citing the target, by
condition (trial-level means over 113 trials per cell). Clustered contrasts
derived from these cells appear in Table~\ref{tab:core}.}
\label{tab:arms}
\end{table}

\section{Cross-Model Replication and Missingness}
\label{app:missingness}

The citation-repair turn is a fixed template parameterized only by the
detected violation (prompt~D.8); it references neither rank nor text arm, so
repair pressure cannot encode the manipulated cues, and repairs are
arm-balanced in the Grok replication.

The Grok~4.3 replication covers 219 of 224 planned cells; five cells missing
after one bounded recovery pass leave 54 of 56 complete pairs. Deterministic
worst-case completion (assigning missing binary outcomes to minimize or
maximize the effect) bounds the 56-pair repaired-protocol structure effect
between $+2.7$ and $+7.1\pp$. Cell-level GPT--Grok agreement: target
incidence 69.9\%, pair citation set 47.0\%, exact pair effect 63.0\%. Grok's
citation-repair rates by condition range from 48.1\% to 61.8\% and are
statistically indistinguishable between arms (structured minus prose repair
probability $+5.6\pp$, 95\% CI $[-4.6,+15.7]$). The attempted 113-pair scaled
Grok replay was aborted under provider rate limiting with seven generated
cells; no partial estimate is reported.

\bigskip
\noindent\textbf{\large Part V \quad Provenance}

\section{Configuration and Model-Visible Interface}
\label{app:config}

\paragraph{Models and decoding.} Acquisition and primary replay: an Azure
OpenAI GPT-5.4 deployment, temperature $0.2$, maximum 2{,}400 completion
tokens, one independently sampled generation per cell (no provider seed
parameter; per-run pseudorandom seeds governing condition order and opaque
handles are archived in every manifest). Evidence review, treatment audit,
cross-model replication, and citation alignment: an Azure Grok-4.3
deployment, temperature $0$--$0.2$ as archived per run.

\paragraph{Model-visible interface.} Each replayed tool message is a JSON
payload with the search query and five result objects containing exactly the
fields \texttt{source\_id}, \texttt{title}, \texttt{url},
\texttt{published\_date}, \texttt{author}, and \texttt{text}. A separate schema
audit inspected 2{,}010 result objects across archived acquisition and replay
payloads; no other field occurs. This audit count differs from the 1{,}750
acquisition result-text records and 1{,}490 deduplicated documents because the
three totals count payload objects, repeated exposures, and unique pages,
respectively.
Provider relevance scores and numeric ranks are never shown; order is
conveyed solely by array position. We verified this directly in the raw
request payloads of \emph{both} the live acquisition and the replay runs
(the archive schema stores Exa's score and rank fields, but the serializer
never includes them in model-visible messages), correcting an erroneous
statement in an earlier draft that scores were shown. Source IDs are deterministic opaque
handles, so identifiers cannot leak the original order.

\paragraph{How the retriever serializes document structure.} The
intervention operates on the serialized \texttt{text} field, downstream of
retrieval and extraction, so what matters is what that field preserves.
Across the 1{,}750 archived acquisition result-text records, 99.8\%
contain newlines, 93.2\% contain markdown-style headings, 76.6\% contain
bulleted or numbered list-marker lines, and 21.3\% contain pipe-delimited
table rows: the Exa text interface retains serialized textual organization
while discarding visual rendering, typography, and DOM structure. Our
manipulation therefore varies a cue class that occurs pervasively in what
the answer model actually reads. Whether publisher-side changes to a live
page would survive retrieval and extraction into this representation is
outside our design
(artifact file \texttt{credit\_supplements.json}).

\paragraph{Citation parsing and repair.} Answers cite
\texttt{[[cite:SOURCE]]} handles, parsed by exact regular-expression match
and mapped back through the per-trial handle table; a citation is valid only
if its handle exists in the transcript. If a final answer contains no valid
citation, the repair turn~D.8 fires once; the scaled GPT runs required zero
repairs.

\paragraph{Treatment lengths.} Original target texts average 1{,}205 words
(median 1{,}367; range 157--1{,}805); polished prose averages 583 (median
482) and polished structured 581 (median 478), reflecting the symmetric
compression both rewrite arms apply; the mechanical-ablation arm reuses the
prose word sequence exactly.

\paragraph{Document and domain recurrence.} The 113 targets span 93 distinct
domains; 11 domains recur across more than one answer-target family (at most
5 families, \texttt{support.google.com}). Re-clustering the primary analysis
by target domain gives $+5.0\pp$, versus $+4.5\pp$ by family.

\section{Archives, Artifact, and Amendments}
\label{app:repro}

Every stage writes a manifest with configuration and content hashes, copied
prompts, append-only records and raw transport traces, parsed citations, and
derived tables. Provider authorization is reserved before each network
request, so retries and failures count against the same declared call budget.

\begin{table}[h]
\centering
\small
\begin{tabular}{@{}p{0.52\columnwidth}l@{}}
\toprule
Stage & Archive \\
\midrule
130-query acquisition & \texttt{20260722-192042} \\
114-pair cohort & \texttt{\ldots\_scale\_v1} \\
113 accepted treatments & \texttt{20260723-035042} \\
452-cell GPT replay & \texttt{20260723-050814} \\
120-cell repeatability run & \texttt{20260723-064207} \\
Grok pilot & \texttt{20260722-041932} \\
Aborted scaled Grok attempt & \texttt{20260723-062625} \\
Mechanical ablation variants & \texttt{20260724-162054} \\
Mechanical ablation replay & \texttt{20260724-162336} \\
Shared-unit alignment & \texttt{20260724-162645} \\
Unit expression/allocation audit & \texttt{20260724-210828} \\
\bottomrule
\end{tabular}
\caption{Scaled-wave archives. The earlier wave's archives are listed in its
own manifests.}
\label{tab:archives}
\end{table}

\paragraph{Protocol amendments.} All are append-only manifest entries with
hashes: fixing the 10{,}000-character Exa text bound; aligning prompt and
executor tool ceilings; replacing lexical top-passage evidence sampling with
contiguous full-exposure chunks (invalidating earlier selective labels);
deterministic exact-quote relocation with conservative downgrade of
unverified quotes; batched evaluator retries with traced backoff; the single
pre-outcome treatment-feasibility exclusion; and the scaled-Grok abort.

\paragraph{Artifact.} The public artifact at
\url{https://github.com/selvamsriram/CiteChoice} contains the full pipeline and
analysis code, all
prompt templates, run configurations, selection rules, derived per-trial and
per-pair tables, summary JSON for every number in this paper, and content
hashes for the frozen inputs. Licensed third-party page
text and raw provider traces are excluded; hashes allow verification against
a licensed re-acquisition. The consolidated analyses
(\texttt{reports/scaled\_paper\_analysis} and
\texttt{reports/revision\_analysis}) regenerate every number with no network
access, and the automated test suite (150 tests) passes.

\end{document}